\documentclass{bmvc2k}

\usepackage{amsfonts}
\usepackage{graphicx}
\usepackage{adjustbox}
\usepackage{amsmath, amssymb}
\usepackage{caption2}
\usepackage{multirow}
\title{Quality Map Fusion for Adversarial Learning}

\addauthor{Uche Osahor}{uo0002@mix.wvu.edu}{1}
\addauthor{Nasser M. Nasrabadi}{ nasser.nasrabadi@mail.wvu.edu}{1}

\addinstitution{
 West Virginia University\\
 Morgantown, USA.
}

\runninghead{Osahor, Nasrabadi}{Quality Map Fusion for Adversarial Learning}

\begin{document}
\maketitle
\begin{abstract}
Generative adversarial models that capture salient low-level features which convey visual information in correlation with the human visual system (HVS) still suffer from perceptible image degradations. The inability to convey such highly informative features can be attributed to mode collapse, convergence failure and vanishing gradients. In this paper, we improve image quality adversarially by introducing a novel quality map fusion technique that harnesses image features similar to the HVS and the perceptual properties of a deep convolutional neural network (DCNN). We extend the widely adopted $l_2$ Wasserstein distance metric  to other  preferable quality norms derived from Banach spaces that capture richer image properties like structure, luminance, contrast and the naturalness of images. We also show that incorporating a perceptual attention mechanism (PAM) that extracts global feature embeddings from the network bottleneck with aggregated perceptual maps derived from standard image quality metrics translate to a better image quality. We also demonstrate impressive performance over other methods.

\vspace{-.4cm}

\end{abstract}

\vspace{-.2cm}
\section{Introduction}
\vspace{-.2cm}
A representation of the human visual system (HVS) is necessary to establish a robust image quality metric which is needed for computer vision applications \cite{10.1117/12.135952,Mohamadi2020DeepBA}. Classical approaches considered hand-crafted strategies to mimic the properties of the HVS  \cite{Wang2004ImageQA, Xue2014GradientMS,Zhang2011FSIMAF, Wang2015AnOB} by implementing a stream of computational functions that are combined to identify the key perceptual properties of images. While these techniques played their role effectively, scaling up these methods have proven to be a daunting task, especially for applications with huge datasets. However, the introduction of neural networks has helped to improve the aforementioned task considerably \cite{Isola2017ImagetoImageTW,Karras2018ProgressiveGO,Osahor2020QualityGS, Kancharla2019QualityAG,Aghdaie2021AttentionAW,  Goodfellow2014GenerativeAN, Mostofa2020JointSRVDNetJS}. These deep networks consist of non-linear filters configured to extract key perceptual features within user-defined constraints from data. 


In our work, we focus more on Full-Reference Image Quality Assessment (FR-IQA) models  \cite{Zhang2012SRSIMAF,Liang2016ImageQA}, these models are mostly used to represent the HVS with the aim of deriving a quality measure for images by comparing the perceptual similarity between distorted images and their respective reference image. A standard FR-IQA model seeks to imitate the HVS by exploiting photographic computational algorithms that represent contrast sensitivity, visual masking, luminance, etc. 
A number of FR-IQA metrics have since being derived which include the Structure Similarity Index Measure (SSIM) \cite{Wang2004ImageQA}, Multi Scale Structure Similarity Index Measure (MS-SSIM)  \cite{10.1007/978-3-319-13671-4_40}, Visual Information Fidelity (VIF) \cite{Han2013ANI}, Feature Similarity Index Measure (FSIM) \cite{Zhang2011FSIMAF}, Mean Deviation Similarity Index (MDSI) \cite{Nafchi2016MeanDS}, etc.
 
In this paper, we present a novel approach for improving  the quality of GAN-synthesised images by combining the benefits of established FR-IQA metrics and the features of a deep convolutional neural network (DCNN). We extend the performance of popular GAN-based baseline approaches by introducing a novel image quality map fusion network that computes the perceptual properties of images and fuse them with a perceptual attention mechanism, as shown in Figure 1. We also introduce novel quality loss functions derived via Banach spaces to boost image quality. Our technique shows impressive results as compared to state-of-the-art. Our key contributions are as follows:
\vspace{-0.1cm}
\begin{itemize}
\item We introduced a new perceptual quality map fusion network that harnesses the perceptual qualities of computationally derived quality assessment metrics.
\vspace{-0.2cm}
\item We propose a new  norm implemented via Banach Wasserstein GAN (BWGAN) instead of the  popular $l_2$ norm computed using the Wasserstein metric.
\vspace{-0.2cm}
\item We also propose a perceptual attention mechanism (PAM) that augments image features to boost the overall visual appeal of the synthesised images.
\end{itemize}
\vspace{-.7cm}
\section{Related Work}
\vspace{-.2cm}
The FR-IQA model tries to simulate the HVS characteristics with good performance measures  \cite{Pei2015ImageQA,Reisenhofer2018AHW,Ye2012UnsupervisedFL,Zhang2014VSIAV}. Two main reasons for the success of FR-IQA can be attributed to the deep learning based perceptual properties of the reference image and the hand-crafted features derived from statistical metrics which are similar to the HVS. Hence, it becomes easier to build a system that minimises the difference between  these two corresponding features. In order to effectively model the properties of the HVS, a couple of related systems have been proposed. Zhang et al.  \cite{Zhang2011FSIMAF} proposed a similarity index metric which calculated the phase congruence and gradient magnitude to represent the HVS system, while  \cite{Xue2014GradientMS} implemented an efficient standard deviation pooling strategy which demonstrated that the gradient magnitude of an image still holds true as a technique for representing the HVS. \cite{Nafchi2016MeanDS} adopted a novel deviation pooling technique to compute the quality score from the gradient and chromaticity similarities as a measure for local structural distortions. 

The  Banach Wasserstein GAN (BWGAN) is a framework that makes use of arbitrary norms other than the popularly used $l_{2}$ norm as the underlying metric of choice in adversarial training. Adler et al. in \cite{Adler2018BanachWG} translated the WGAN-GP model to Banach spaces which have the capacity to utilize norms that capture desired image features like edges, texture, etc.
\vspace{-.1cm}

\vspace{-.2cm}
\begin{figure*}
\begin{center}
  \includegraphics[width=12.8cm, height=5cm ]{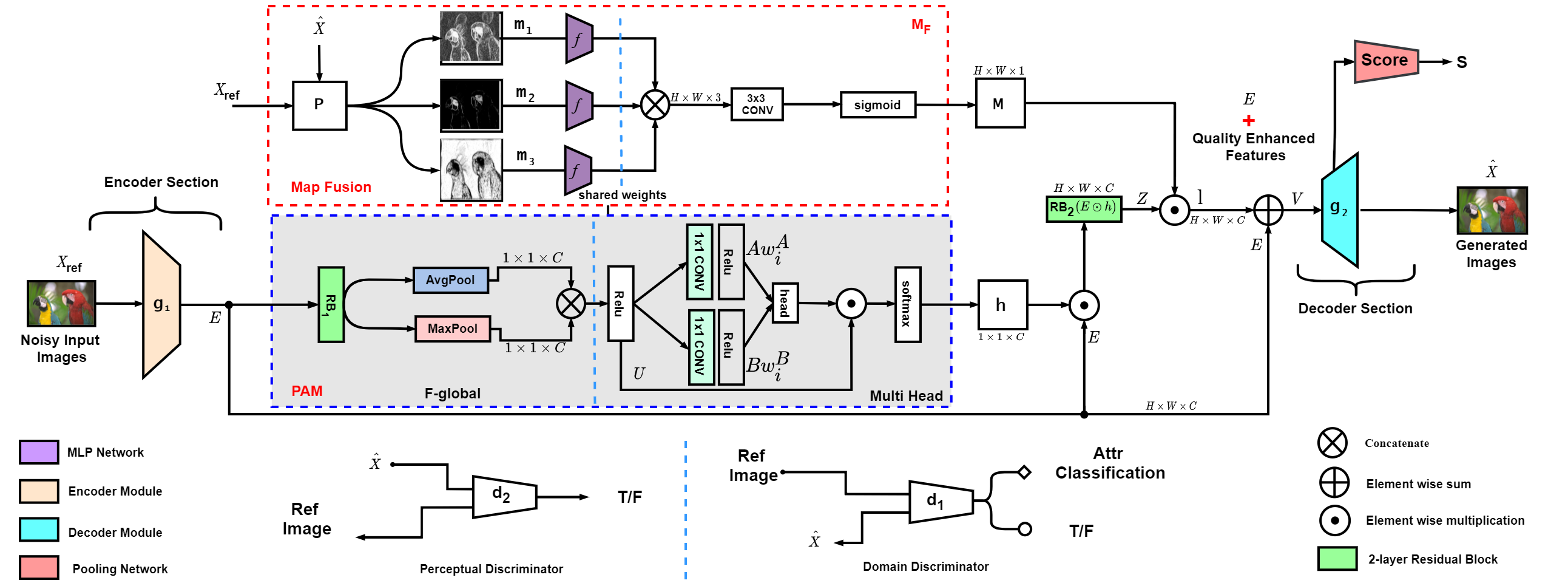}
\end{center}
   \caption{The quality encoding architecture. The structure shows the  generator $G:(g_1, g_2)$ configuration coupled with a quality Map Fusion network $M_{F}$. The domain discriminator $d_1$ (lower-right) extracts features where True/False predictions are made per pixel and  attribute classification is executed to ensure multi-domain adaptation, $d_2$ critics the perceptual map to maintain consistency in an adversarial manner.}
\label{fig:long}
\label{fig:onecol}
\vspace{-.3cm}
\end{figure*}

\section{Our Approach}
\vspace{-.15cm}
We present a single  GAN model capable of implementing image-to-image synthesis. We combine the benefits of established FR-IQA metrics \cite{Zhang2011FSIMAF, 10.1007/978-3-319-13671-4_40, Nafchi2016MeanDS} and the low-level salient features of a deep convolutional neural network (DCNN) to  aid adversarial image synthesis aimed at producing perceptually appealing images. In our model, we introduced an attention schema that exploits the salient perceptual features in a channel-wise fashion and the spatial map representation embeddings of standard FR-IQA metrics (SSIM, MDSI and FSIM). Our framework consists of five main components; a quality-aware generator network ${G:(g_1, g_2)}$, where $g_1$ and $g_2$ represent the encoder and decoder section, respectively.
$g_1$ and $g_2$ are coupled with a perceptual attentive mechanism (PAM) for quality encoding and  a  perceptual quality map fusion network ${M_F}$  at the latent space as shown in Figure 1. The discriminative networks ${D:(d_1, d_2)}$ critics images generated by ${G:(g_1, g_2)}$ in an adversarial manner without compromising image quality and the perceptual consistency with the reference image.  The perceptual quality map generator combines the core quality metric functions that capture the sensitive perceptual features of a given image, while the score regression network pools the images synthesized by ${G:(g_1, g_2)}$ to estimate reference quality score $s$. Our  overall objective consists of a Wasserstein Gradient Penalty, Structural Similarity Index Gradient Penalty (SSIM-GP) and a Natural Image Quality Estimator (NIQE) as defined in section 4.

\vspace{-.2cm}
\subsection{Perceptual Attention Mechanism (PAM)}
\vspace{-.2cm}
The attention mechanism augments perceptual features from a prior generator encoder network $g_1(\cdot)$ computed over input images $X_i$. The aim is to establish a convex combination of quality-enhanced condensed representations of the input image for real time training. We begin by describing the channel attention in PAM, which is based on the CBAM module \cite{Woo2018CBAMCB}. PAM involves two steps:  first, per-channel “summary statistics” $F_{global}$ obtained from a 2-layer residual block $RB_1$, is calculated to yield the global feature attention vector $ U \in \mathbb{R} ^{1 \times 1 \times C}$. Secondly,  a multi-head network  $head_{i} = Attention(Aw_i^{A},Bw_i^{B} )$ applies a non-linear  multi-head attention transformation which allows the model to jointly attend to information $A$ and $B$ from different representation sub-spaces $w_i^{A}$ and $w_i^{B}$ of the bottleneck \cite{Vaswani2017AttentionIA}. The channel-based attention output is given as $h$ = softmax $(head_i \odot U)$ which is multiplied with the encoder output $E \in \mathbb{R}^{H \times W \times C}$ from $g_1$ and processed by the residual block $RB_2$ to produce the channel-based attention embeddings, denoted as $ Z = RB_2( E \odot{h})$ where $ Z \in \mathbb{R}^{H \times W \times C}$. 
\vspace{-.2cm}
\subsection{Perceptual Map Generator}
\vspace{-.2cm}
We selected the FSIM \cite{Zhang2011FSIMAF}, MS-SSIM \cite{10.1007/978-3-319-13671-4_40}, and MDSI \cite{Nafchi2016MeanDS} image quality metrics to generate similarity maps because the trio collectively capture key image characteristics that are similar to the HVS \cite{Wang2004ImageQA,Nafchi2016MeanDS,10.1007/978-3-319-13671-4_40,Zhang2011FSIMAF,Aghdaie2021DetectionOM} as shown in Figure 2. The FSIM metric captures the luminance, contrast and structural information. For the MS-SSIM metric, we considered multiple scales of the synthesised image and its reference for contrast and structure while the MDSI map is derived by extracting the gradient and chromaticity of the  pair of  synthesised and reference images, respectively. We use an intensity coefficient; $ 0.3 \leq \alpha \leq 1$ to specify the intensity of the maps.
The map fusion network $M_F$ is divided into three stages. First, we extract the feature similarity representations between the reference image $X_{ref}$ and the generated images $\hat{X}_i$ per iteration given as  $ p(X_{ref}, \hat{X}_i)$  for the aforementioned similarity metrics, where $p(\cdot)$ is an arbitrary function used to calculate similarity index maps; MS-SSIM, FSIM and MDSI. Secondly, the generated maps, ($m_1, m_2, $ and $m_3$) are concatenated and pre-processed by two-layer MLP networks $f(\cdot)$ to form a spatial-based perceptual map representation $M \in \mathbb{R}^{H \times W \times 1}$. 
At the last stage, the predicted future states $I$ is then computed as the expectation of spatial features $M$ and  the channel-based features  $ Z \in \mathbb{R}^{H \times W \times C}$. 
$I$ is then summed with the output of the encoder $E$ given as $V=I+ E$. The resulting output $ V \in \mathbb{R}^{H \times W \times C}$ which is fed to decoder $g_2$ represents latent features  that are optimized for better image quality.

\begin{figure*}
\begin{center}
\includegraphics[width=12.7cm, height=2.8cm]{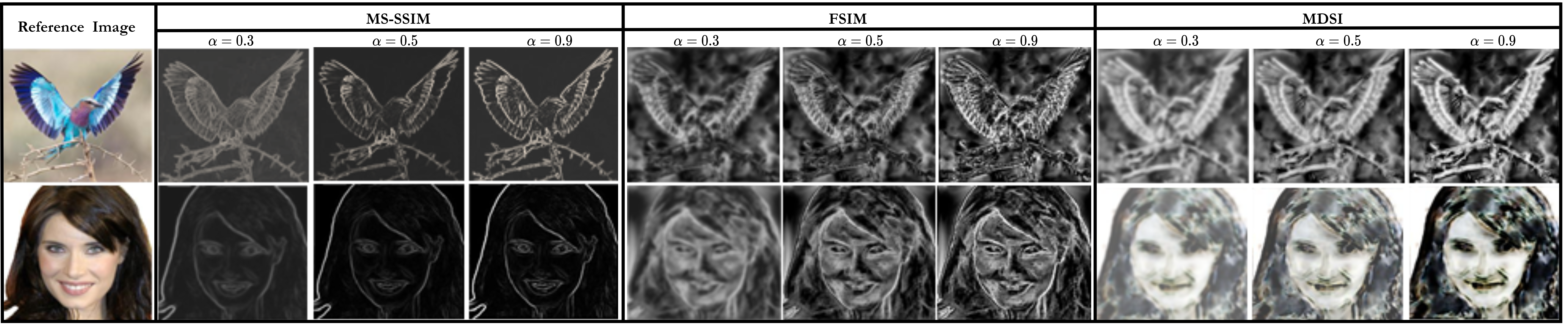}
\end{center}
\vspace{-.3cm}
   \caption{A spectrum of quality maps obtained at different intensities. 
   }
\label{fig:attribute}
\vspace{-.4cm}
\end{figure*}

\vspace{-.3cm}
\section{Banach space gradient penalty}
\vspace{-.2cm}
Quality assessment metrics for the distance between  images has been limited to cost functions that take the form of $l_1$ or $l_ 2$ norms. However, issues like  non-convexity and complications in gradient computations (vanishing gradients, exploding gradients, etc) are some of the struggles experienced in formulating optimization problems. 
To mitigate the aforementioned computational shortcomings, the Wasserstein distance was introduced in  \cite{Arjovsky2017WassersteinG}.
However, a wide variety of untapped  metrics \cite{10.1007/978-3-319-13671-4_40, Wang2004ImageQA} exist that can be used to compare and emphasize key features of interest. 
In this regard, we extended the Wasserstein distance  beyond the popular WGAN with the gradient penalty (WGAN-GP), which is constrained to $l_2$ norms and rather adapted a more complete space called the Banach space \cite{Adler2018BanachWG}. Our technique, similar to  \cite{Arjovsky2017WassersteinG, Kancharla2019QualityAG, Adler2018BanachWG} shows that the characterisation of $\gamma$-lipschitz functions via the norm of the differential can be extended from the $l_2$ setting to arbitrary Banach spaces by considering the gradient as an element in the dual of $B$. 
Such a loss function is given as: 
\begin{equation}
\begin{aligned}
L_{B} =\lambda E_{Y}\bigg( \frac{1}{\gamma}   \left\| \partial D(Y)\right\|_B^* - 1\bigg )^2 ,
  \end{aligned}
\end{equation}
where $\lambda, \gamma \in \mathbb{R}$, are regularization parameters. These Banach space norms give room for specific image  features such as texture, structure, contrast and luminance which highlight the perceptual appeal of a  human observer, as described in section 4.1 and 4.2.
\vspace{-.2cm}
\subsection{Structural Similarity (SSIM) index}
The SSIM index measures the perceptual difference between two similar images. The local mean, variance and structure are computed to find an local quality score \cite{Wang2004ImageQA}. The SSIM index computes changes to local mean, local variance and local structure between two images $X$ and $\hat{X}$. The local scores  are then averaged across the image to find the image quality score. 

\vspace{-0.4cm}
\begin{equation}
\begin{aligned}
     L(X_{(i,j)},{\hat{X}}_{(i,j})  = \frac{2 \mu_{x}(i,j) \mu_{\hat{x}}(i,j) + C_1}{\mu^{2}_{x}(i,j) + \mu^{2}_{\hat{x}}(i,j) + C_1} 
    ,\ CS(X_{(i,j)},{\hat{X}}_{(i,j}) = \frac{2 \sigma_{x\hat{x}}(i,j) + C_2}{\sigma_{x^{2}}(i,j) + \sigma_{\hat{x}^{2}}(i,j) + C_2},  
\end{aligned}
\end{equation}
\begin{equation}
\begin{aligned}
         & d_{b} (X_{(i,j)},{\hat{X}}_{(i,j}) = \sqrt{2 - L((X_{(i,j)},{\hat{X}}_{(i,j})) - CS(X_{(i,j)},{\hat{X}}_{(i,j)}}),  
    \end{aligned}
\end{equation}
\noindent
where $X$ and  $\hat{X}$ refer to the input and synthesised images, the subscript $(i,j)$ is the pixel index, $\mu_{(i,j)}$ and $\sigma_{(i,j)}$ are the local mean and standard deviation, respectively. $L(X_{(i,j)},\hat{X}_{(i,j)})$, $C(X_{(i,j)},\hat{X}_{(i,j)})$ and  $S(X_{(i,j)},\hat{X}_{(i,j)})$ are the local luminance, contrast and structure scores at pixel $(i,j)$, respectively. Furthermore, since $d_b(X,\hat{X})$ is bounded, the lipschitz constant can be imposed directly by introducing a gradient penalty regularization term given as:
\begin{equation}
\begin{aligned}
        SSIM GP = E_{X \sim P_{X}, \hat{X} \sim P_{\hat{X}}} 
        \bigg[\bigg( \frac{|D(X) - D(\hat{X})|}{d_b(X,\hat{X})}     \bigg ) -1\bigg]^{2}.
    \end{aligned}
\end{equation}
This makes the SSIM a good candidate for quality awareness which is beneficial for regularizing GANs. The complete mathematical properties are described in \cite{Brunet2012OnTM}.

\begin{figure*}
\begin{center}
\includegraphics[width=12.2cm, height=3.9cm]{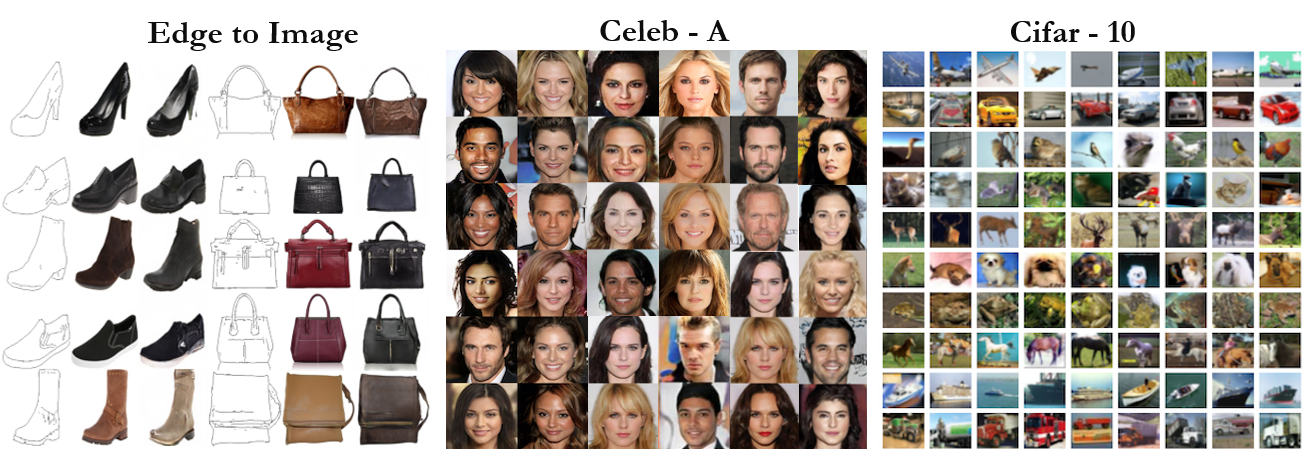}
\end{center}
\vspace{-.3cm}
   \caption{A sample of synthesised images representing different datasets.
   }
\label{fig:attribute}
\vspace{-0.4cm}
\end{figure*}
\vspace{-.2cm}
\subsection{Natural Image Quality Estimator (NIQE)}
The NIQE \cite{Mittal2013MakingA} is an NR-IQA metric of perceptually  relevant  spatial  domain  Natural Scene Statistics (NSS)  features  extracted  from local image patches that capture the essential low-order  statistics  of  natural  images. The equation is given as:

\begin{equation}
\begin{aligned}
         &  \hat I{(i,j)} = \frac{I_{(i,j)} - \mu_{(i,j)}}{\sigma_{(i,j)} + 1}, 
    \end{aligned}
\end{equation}
\noindent
where $ I_{(i,j)} $ is the pixel index and $\mu_{(i,j)} $ and $ \sigma_{(i,j)}$ are the local mean and standard deviation. The NIQE captures the naturalness of a pristine reference image  by modelling a generalized gaussian distribution (GGD) \cite{Ruderman1994TheSO}, and models the products of neighbouring pixel coefficients using an Asymmetric GGD (AGGD). The parameters of both the GGD and AGGD are then modelled using a Multivariate Gaussian Model (MVG) distribution \cite{Moorthy2010StatisticsON}. The quality of the test image is measured in terms of the “distance” of its MVG parameters $\mu_{t} $ and $ \sigma_{t}$ from the pristine MVG parameters obtained. Finally, discriminator gradients computed for both pristine reference  and synthesised images are used to compute the distance between the pair. 
The expression is given as:
\vspace{-0.1cm}
\begin{equation}
\vspace{-0.2cm}
\begin{aligned}
         &  \left\|  (\mu_{X}, \Sigma_{X} )\right\|_{NIQE}  := \sqrt{(\mu_{X} - \mu_{\hat{X}})^{T} \bigg ( \frac{\Sigma_{X} + \Sigma_{\hat{X}}}{2}\bigg )^{-1}(\mu_{X} - \mu_{\hat{X}}),} 
    \end{aligned}
\end{equation}
\noindent
where $\mu_{X}$, $\mu_{\hat{X}}$,  $\Sigma_{X}$ and $\Sigma_{\hat{X}}$  are the  mean and covariance of the reference $X$ and synthesised $\hat{X}$ images, respectively.
In addition to the SSIM and NIQE metrics, we also used a 1-GP regularizer \cite{Gulrajani2017ImprovedTO} designed to force the local statistics of the discriminator gradient to be as close to those of real images. Our claim is that such a regularization strategy results in improving visual quality of the generated images especially for attributes like hair, age, skin colour etc. We worked in the WGAN-GP framework to demonstrate our method. The overall discriminator cost function includes the NIQE function regularizer, the SSIM and the 1-GP regularizer defined as:
\vspace{-0.1cm}
\begin{equation}
\vspace{-0.4cm}
\begin{aligned}
  & L_{BP} =  \lambda_{1} E_{\hat {x}\sim P_{\hat{x}}} (\left\| \nabla_{\hat{x}} D(\hat{X}) | \mu_{P}, \Sigma_{P}\right\|_{NIQE}) + 
    \lambda_{2} E_{x \sim P_{x}, \hat{x} \sim P_{\hat{x}}} 
        \bigg[\bigg( \frac{|D(X) - D(\hat{X})|}{d_b(X,\hat{X})}     \bigg ) -1\bigg]^{2}_{SSIM} & \\
        \; &   \hspace{3cm} +  \lambda_{3} E_{x\sim P_{\hat{x}}} (\left\| \nabla_{\hat{x}} D(\hat{X}) |\right\|_{1-GP}),
   \end{aligned}
\end{equation}
\noindent
The full objective is given as:
\begin{equation}
\begin{aligned}
  {L}_{GAN}(G,D,X,\hat{X})  &= {E}[log D(X)) + 
  {E}[log (1- D(G(\hat{X})) - s)] + L_{BP} ,
  \end{aligned} 
  \end{equation}
\noindent
where $s$ is the generated score from the regression network minimised over the groudtruth scores of the images.
we use $\lambda_{1}, \lambda_{2} $ and $ \lambda_{3}$ as a means of tuning the objective functions to achieve better results. 

\begin{figure}
\hspace{0.30cm}
\begin{minipage}[b]{0.30\linewidth}
\centering
\includegraphics[height=2.9cm, width=3.8cm]{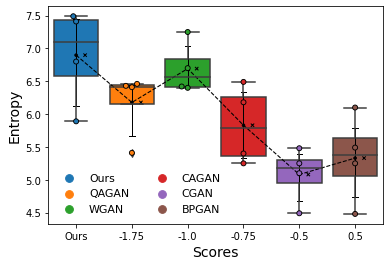}
\captionlabelfalse 
\caption{(a) Entropy}
\end{minipage}
\begin{minipage}[b]{0.30\linewidth}
\centering
\includegraphics[height=2.9cm, width=3.8cm]{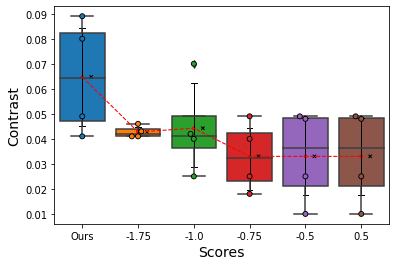}
\captionlabelfalse 
\caption{(b) Contrast}
\label{fig:figure2}
\end{minipage}
\begin{minipage}[b]{0.30\linewidth}
\centering
\includegraphics[height=2.9cm, width=3.8cm]{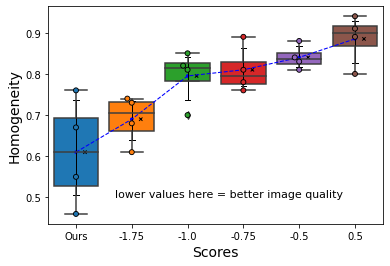}
\captionlabelfalse 
\caption{(c) Homogeneity}
\label{fig:figure2}
\end{minipage}
\vspace{0.2cm}
\captionlabelfalse 
\vspace{0.1cm}
\caption {Figure 4: Statistical feature IQA metric values. }
\vspace{-0.2cm}
\end{figure}

\begin{figure}[t]
 \hspace{0.2cm}
\begin{minipage}[b]{5cm}
\centering
\includegraphics[height=3.1cm, width=6cm]{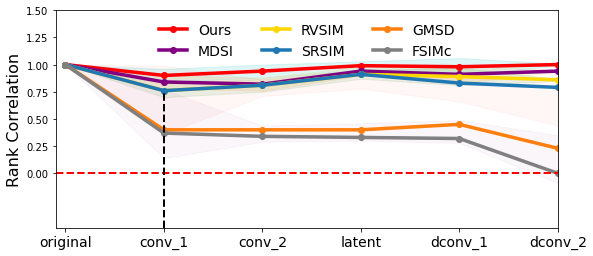}
\captionlabelfalse 
\caption{(a) FairFace Dataset}
\label{fig:figure2}
\end{minipage}
 \hspace{0.9cm}
\begin{minipage}[b]{5cm}
\centering
\includegraphics[height=3.1cm, width=6cm]{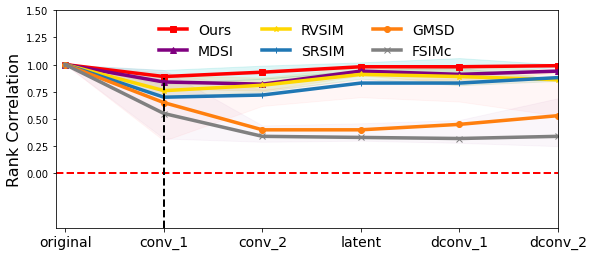}
\captionlabelfalse 
\caption{(b) CelebA Dataset}
\label{fig:figure2}
\end{minipage}
\vspace{0.2cm}
\captionlabelfalse 
\caption{Figure 5: Superman's rank correlation values at different layers of the network}
\vspace{-.4cm}
\end{figure}

\vspace{-.2cm}
\begin{figure*}[t]
\begin{center}
\includegraphics[height=3cm, width=11.7cm]{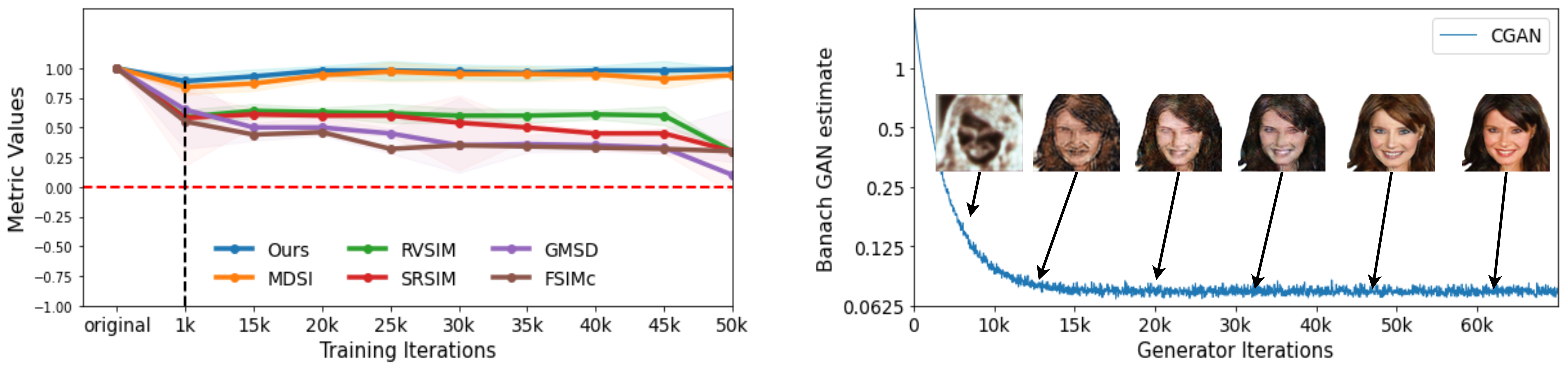}
\vspace{-0.20cm}
\end{center}
\captionlabelfalse 
\caption{(a) \hspace{5cm} (b)}
\vspace{0.20cm}
\captionlabelfalse
\caption{Figure 6: HOG similarity metrics (left) and synthesised image results for FairFace Dataset.}
\vspace{-0.1cm}
\end{figure*}

\vspace{-.2cm}
\begin{figure*}[t]
\begin{center}
\includegraphics[width=11.7cm, height=3.4cm]{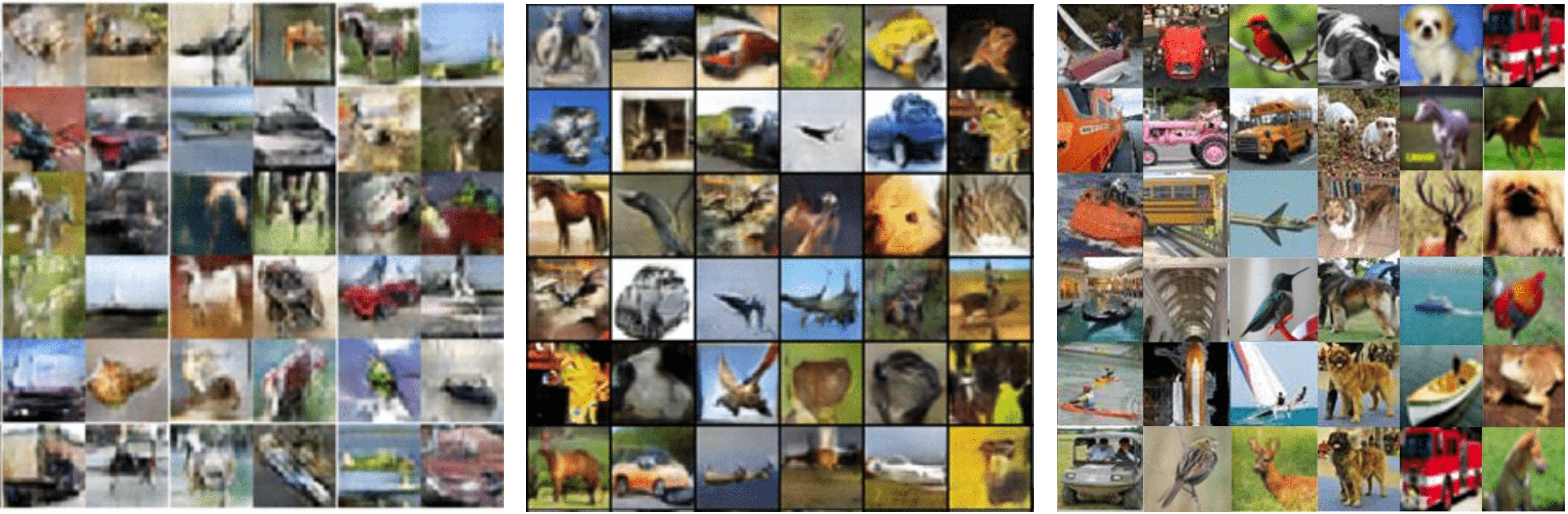}
\end{center}
\vspace{-0.15cm}
\captionlabelfalse 
\vspace{0.10cm}
\caption{(a) \hspace{3cm} (b) \hspace{3cm} (c)}
\vspace{0.2cm}
\captionlabelfalse
  \caption{Figure 7: Randomly sampled images generated using a combination of our model base line ($BL$) and different losses for W-GAN and BWGAN (SSIM and NIQE) of the CIFAR-10 dataset. 7 (a) Shows images synthesised using $BL$ and just the WGAN with gradient penalty loss. 7 (b) performs better when the SSIM loss function is added ($BL+WGAN+SSIM$). 7 (c) shows the best results when all loss functions are included ($BL+WGAN+SSIM+NIQE$).
}
\label{fig:attribute}
\vspace{-.3cm}
\end{figure*}

\vspace{-.2cm}
\begin{figure*}
\begin{center}
\includegraphics[width=11.7cm, height=3.4cm]{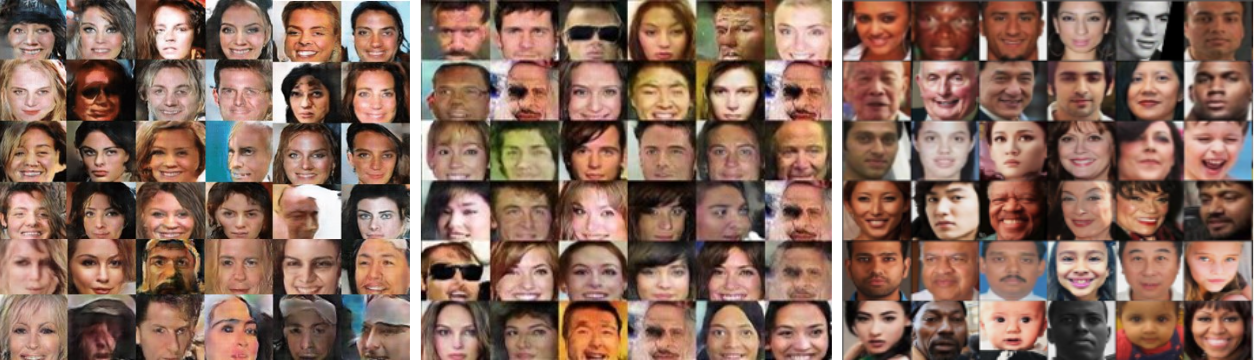}
\end{center}
\vspace{-0.1cm}
\captionlabelfalse 
\caption{(a) \hspace{3cm} (b) \hspace{3cm} (c)}
\vspace{0.2cm}
\captionlabelfalse
   \caption{ Figure 8: Randomly sampled images generated using different models for FairFace dataset. 8 (a) Shows images synthesised using DCGAN. 8 (b) performs better due to the gradient penalty approach of WGAN. However, 8 (c) shows that our model performs even better when Banach losses are included.
   }
\label{fig:attribute}
\vspace{0.2cm}
\end{figure*}

\vspace{0.1cm}
\section{Training Strategy}
\vspace{-.2cm}
We trained our model using the Adam optimizer, with momentum values set at  $\beta_1 $ = 0.5 and  $\beta_2$  = 0.99, we used a batch size of 8 for most experiments on CelebA  \cite{liu2015faceattributes}, Celeba-HQ  \cite{Lee2019MaskGANTD} and FairFAce  \cite{Krkkinen2019FairFaceFA}, respectively. A learning rate of 0.0001 for the first 10 epochs which linearly decayed to 0 over the corresponding epochs.  We trained the entire model on  three NVIDIA Titan X GPUs.

\vspace{-.3cm}
\subsection{Datasets}
\vspace{-.2cm}
We evaluated the efficacy of our proposed technique on the following datasets:
The CelebFaces Attributes (CelebA) \cite{liu2015faceattributes} of 202,599 celebrity face images. We cropped the initial images to 178x178, then resized them to 64x64. The CIFAR-10 \cite{Krizhevsky09learningmultiple} dataset consists of 60,000 32x32 colour images in 10 classes, with 6,000 images per class. The Fair-Face \cite{Krkkinen2021FairFaceFA} image dataset contains 108,501 images, with an emphasis of balanced race composition in the dataset comprising 7 race groups: White, African, Indian, East Asian, Southeast Asian, Middle East, and Hispanic. For evaluations, we used the LIVE \cite{Sheikh2006ASE} which consists of 982 distorted images with 5 different distortions. The TID2008 dataset \cite{Reisenhofer2018AHW} that contains 25 reference images and a total of 1,700 distorted images. We also used Edges-to-shoes 50,000 training images from UT Zappos50K dataset \cite{Yu2014FineGrainedVC}  and Edges-to-Handbag 137,000 Amazon Handbag images from \cite{Zhu2016GenerativeVM}, trained for 15 epochs and batch size 8. 
\vspace{-.20cm}
\subsection{Evaluation}
\vspace{-.10cm}
To evaluate the performance of the synthesized images as shown in Figure 3,
two key evaluation criteria were adopted in our paper; the Spearman’s Rank Order Correlation Coefficient (SROCC) and the Linear Correlation Coefficient (LCC) \cite{Forthofer1981}.
SROCC is a measure of the monotonic relationship between the ground-truth and model prediction, while the LCC is a measure of the linear correlation between the ground-truth and model prediction. Table 1 and 2 shows the SROCC and LCC performance of the competing IQA methods for different distortion types, respectively.
In general, our model  performs competitively among most distortion types. Compared with BPSQM, our model shows more performance of about 4.5\% overall in dealing with the distortion of AGN, SCN, HFN, JPEG and MN, respectively as indicated in Table 1. For comparison with previous models, we computed three quantitative measures: Inception Score (IS), Frechet Inception Distance (FID) and the Feature Similarity (FSIM) index.  IS measures the sample quality and diversity by finding the entropy of the predicted labels. FID score measures the similarity between real and fake samples by fitting a multivariate Gaussian (MVG) model to the intermediate representation. The FSIM index computes quality estimates based on phase congruency as the primary feature, and incorporates the gradient magnitude as the complementary feature for the real and fake samples, respectively. Table 3 shows the quantitative comparison of the GAN-metric performance for BPGAN \cite{Wang2017BackPA}, CAGAN \cite{Yu2017TowardsRF}, CGAN \cite{CycleGAN2017} , WAGAN \cite{Arjovsky2017WassersteinG}, QAGAN \cite{Kancharla2019QualityAG} and ours for CelebA dataset.
\vspace{-0.1cm}
We also carried out pixel variation analysis on the synthesised images by using the second order features of the synthesized images, which are based on the gray level co-occurrence matrix (GLCM) \cite{Haralick1979StatisticalAS}.  We used the aforementioned technique to determine  the Entropy, Homogeneity and Correlation of the synthesised images in comparison with state-of-the-art methods as shown in Figure 4.  Entropy is useful for assessing sharpness while Homogeneity and Correlation are useful for evaluating the Contrast of an image.  Entropy and Correlation increase in image quality, whereas Homogeneity energy values decrease with increase in image quality. From the Entropy plot in 5(a), our model performs decently well by over 3.5\% compared to the QAGAN and WAGAN. The Contrast level improves drastically for our approach as compared to the other methods that are closely matched at a tolerance of about 2\%. We also observed that most models possess similar homogeneity values except our model and QAGAN which reflect significant performance values.

\begin{table}
\vspace{.3cm}
\begin{center}
\caption{
SROCC comparison on individual distortion types on the TID2008 databases.
}
\begin{adjustbox}{height = 1.6cm, width=12.0cm}
\begin{tabular}{c|cccccccccccccc} 
\hline
{SROCC} & \multicolumn{12}{c}{TID2008} \\
\cline{2-14} & AGN & ANMC & SCN &  MN & HFN &  IMN &  QN &  GB &  DEN &  JPEG & JP2K & JGTE & J2TE \\ 
\hline
GMSD \cite{Xue2014GradientMS} & 0.911 & \textbf{0.888} & 0.914 & 0.747 & 0.919 & 0.683 & 0.857 & 0.911 & \textbf{0.966} & 0.954 & 0.983 & 0.852 & 0.873\\
FSIMc \cite{Zhang2011FSIMAF} & 0.910 & 0.864 & 0.890 & 0.863 & 0.921 & 0.736 & 0.865 & 0.949 & 0.964 & 0.945 & 0.977 & 0.878 & \textbf{0.884}\\
BLIINDSII \cite{Yang2018BlindIQ}  & 0.779 & 0.807 & 0.887 & 0.691 & 0.917 & 0.908 & 0.851 & 0.952 & 0.908 & 0.928 & 0.940 & 0.865 & 0.855\\
DIIVINE \cite{Moorthy2011BlindIQ}  & 0.812 & 0.844 & 0.854 & 0.713 & 0.922 & 0.915 & 0.874 & 0.943 & 0.912 & 0.930 & 0.938 & 0.873 & 0.852\\
BRISQUE \cite{Mittal2012NoReferenceIQ} & 0.853 & 0.861 & 0.885 & 0.810 & 0.931 & \textbf{0.927} & 0.881 & 0.933 & 0.924 & 0.934 & 0.944 & \textbf{0.891} & 0.836\\
NIQE \cite{Mittal2013MakingA} & 0.786 & 0.832 & 0.903 &  0.835 & 0.931 & 0.913 & 0.893 & \textbf{0.953} & 0.917 & 0.943 & 0.956 & 0.862 & 0.827\\
BPSQM \cite{Pan2018BlindPS} & 0.881 & 0.801 & 0.935 & 0.786 & 0.938 & \textbf{0.933} & \textbf{0.920} & 0.937 & 0.914 & 0.943 & 0.967 & 0.829 & 0.644\\
\hline
Ours & \textbf{0.936} & 0.878 & \textbf{0.961} & \textbf{0.939} & \textbf{0.948} & 0.892 & 0.915 & 0.898 & 0.878 & \textbf{0.955} & \textbf{0.987} & 0.836 & 0.779
\\
\hline
\end{tabular}
\end{adjustbox}
\end{center}
\vspace{-0.45cm}
\end{table}
\begin{table}[t]
\begin{minipage}[b]{0.50\linewidth}
\centering
\vspace{0.3cm}
   \caption{
LCC evaluation on  LIVE database.
}
\begin{adjustbox}{height = 1.5cm, width=5.9cm}
\begin{tabular}{c|ccccccc} 
\hline
{LCC} & \multicolumn{5}{c}{LIVE} \\\\
\cline{2-7} & JP2K & JPEG & WN & BLUR & FF & ALL\\ 
\hline
BRISQUE \cite{Mittal2012NoReferenceIQ} & 0.923 &0.973 &0.985& 0.951 &0.903& 0.942\\
CORNIA \cite{Ye2012UnsupervisedFL} &0.951 &0.965& 0.987& 0.968& 0.917 & 0.935\\
CNN \cite{Kang2014ConvolutionalNN} & 0.953 & 0.981 & 0.984 & 0.953 & 0.933 & 0.953\\
SOM \cite{Zhang2015SOMSO} & 0.952 & 0.961 & 0.991 & 0.974 & 0.954 & 0.962\\
BIECON \cite{Kim2017FullyDB} & 0.965 & \textbf{0.987} &  0.970 & 0.945 & 0.931 & 0.962\\
\hline
Ours & \textbf{0.975} & 0.986 & \textbf{0.994} & \textbf{0.988} & \textbf{0.960} & \textbf{0.982}\\
\hline
\end{tabular}
\end{adjustbox}

\label{table:student}
\end{minipage}
\begin{minipage}[b]{0.50\linewidth}
\centering
\vspace{0.3cm}
    \caption{ GAN-metric performance.}
    \label{table:student}
\begin{adjustbox}{height = 1.5cm, width=3.5cm}
\begin{tabular}{c|cccc} 
\hline
{ Model} & \multicolumn{2}{c}{CelebA} \\\\ 
\cline{2-4} & FID $\downarrow$ & FSIM $\uparrow$  & IS $\uparrow$ \\ 
\hline
BPGAN \cite{Wang2017BackPA}    & 86.10   &  69.13  &  0.87 \\ 
CGAN \cite{CycleGAN2017} & 43.21   &  71.10  &  0.89 \\
CAGAN \cite{Yu2017TowardsRF}    & 36.16   &  71.33   &  0.90 \\
WGAN \cite{Arjovsky2017WassersteinG}   & 33.24   &  72.60   &  0.91 \\
QAGAN \cite{Kancharla2019QualityAG}   & {18.23}   &  82.69   &  0.96 \\
\hline
Ours & \textbf{18.39 }  &  \textbf{83.40 } &  \textbf{0.97} \\
\hline
\end{tabular}
\end{adjustbox}
\end{minipage}
\vspace{-0.3cm}
\end{table}
\subsection{Ablation Study}
\vspace{-0.15cm} 
Ablation studies on our loss functions was implemented to test model robustness in general for the CIFAR-10, FairFace and CelebA datasets, respectively. The Lagrange coefficients $\lambda_{1} $ and $ \lambda_{2} $ of the SSIM and NIQE losses were also changed empirically within ($ 0.001 \geq\lambda_{1} $ and $ \lambda_{21}\leq 1.000$) range, to check the effect on the perceptual appeal of the synthesised images. It was inferred that reducing the coefficients towards the lower limit weakens the discriminative power which in turn reduces the quality of the synthesised images from the generator. 
We also conducted a Histogram of Oriented Gradient (HOG) similarity performance with the Inception v3 model \cite{Szegedy2016RethinkingTI} for the input and synthesised images on the FairFace dataset, in order  to obtain the model layer-wise performance at specific iterations of the baseline of our model. Figure 6 (a) shows the HOG similarity performance at different iterations while training for our model compared to other quality metric techniques. 

Our results show that our approach is closest to MDSI \cite{Nafchi2016MeanDS}, as compared to RVSIM \cite{Ye2012UnsupervisedFL}, GSMD \cite{Xue2014GradientMS}, SRSIM \cite{Zhang2012SRSIMAF}, FSIMc \cite{Zhang2011FSIMAF} that perform slightly below our model. An SRCC plot representation in Figure 5 depicts the rank correlation performance for both FairFace and CelebA dataset. The values confirm that our model performs favourably over other aforementioned techniques. At different iteration values, we also observed decent image quality improvements at about 20k - 30k iterations as shown in Figure 6(b) for the FairFace dataset. Figures 7 and 8 show further results obtained from a combination of different loss functions and other competitive models, respectively. 

We computed the FID and IS scores of synthesised images for ClebeA and CIFAR-10 datasets with resolutions of 64 x 64 and 32 x 32, respectively. Table 4 shows the performance of our model baseline (BL) for different combinations of attention schemes (PAM and $M_F$) and the IQA losses (NIQE and SSIM). By observation, we see from Table 4 that including  the $M_F$ module significantly boost image quality, this is a confirmation that perceptual spatial salient maps are crucial in GAN models for better image quality \cite{Yang2018BlindIQ, Saad2012BlindIQ, Johnson2016PerceptualLF}. 

Furthermore, we applied the PAM and $M_F$ attention modules to the  StleGAN2 \cite{Karras2020AnalyzingAI}  architecture. We also added the proposed Banach space norms (SSIM NIQE) to compare the overall model performance with our model. In Table 5, we show the trade-offs of the QAGAN \cite{Kancharla2019QualityAG}, StyleGAN2 \cite{Karras2020AnalyzingAI} and our model baseline (BL). We used different combinations of the standard IQA metrics as discussed in section 4.1 and 4.2. Our findings confirm that our approach is competitive with state-of-the-art. Most importantly, we see improved performance of  our model  at lower resolutions (32 x 32), this improvement can be attributed to the attention schema employed. In Figure 9, we showcase the performance of QAGAN \cite{Kancharla2019QualityAG} and our model on image synthesis for CelebA dataset. Our results show that our model performs significantly well overall, Table 5 gives a clearer representation of the performance levels.
\begin{table}
\vspace{-0.3cm}
\centering
\vspace{0.6cm}
    \caption{ Ablation study on CelebA and CIFAR-10 datasets on our model baseline "BL" with a combination of the quality modules "PAM" and "$M_F$" and losses "SSIM" and "NIQE".}
    \vspace{0.2cm}
\begin{adjustbox}{height = 1.5cm, width = 12cm}
\begin{tabular}{lccccc|cccc} 
\hline
& \multirow{2}{*} & & & & & \multicolumn{2}{c}{\textbf{CIFAR-10}} &  \multicolumn{2}{c}{\textbf{CelebA}} \\
\cline{7-10} & \textbf{BL} & \textbf{PAM} & \textbf{M}{$_F$} &  \textbf{NIQE} & \textbf{SSIM}  & \textbf{FID}$\downarrow$ &   \textbf{IS} $\uparrow$ & \textbf{FID}$\downarrow$ &  \textbf{IS} $\uparrow$ \\
\hline
& \centering\checkmark &  &  &  &  & 38.10 $\pm$ 0.12 & 8.20 $\pm$ 0.03 & 29.80 $\pm$ 0.10 & 0.86 $\pm$ 0.03 \\ 
& \centering\checkmark  & \centering\checkmark &  &  & \centering\checkmark & 19.01 $\pm$ 0.10 & 8.01 $\pm$ 0.13 & 13.16 $\pm$ 0.02 & 0.88 $\pm$ 0.19 \\
& \centering\checkmark  & \centering\checkmark &  &\centering\checkmark  &  & 16.31 $\pm$ 0.21 & 8.00 $\pm$ 0.35 & 11.86 $\pm$ 0.07 & 0.89 $\pm$ 0.11 \\
& \centering\checkmark & & \centering\checkmark & & \centering\checkmark & 15.00 $\pm$ 0.20 & \textbf{7.46 $\pm$ 0.21} & 10.76 $\pm$ 0.43 & 0.90 $\pm$ 0.10 \\
& \centering\checkmark & & \centering\checkmark & \centering\checkmark &  & 13.21 $\pm$ 0.10 & 7.80 $\pm$ 0.10 & \textbf{6.38 $\pm$ 0.39} & 0.96 $\pm$ 0.10 \\
&\centering\checkmark  & \centering\checkmark & \centering\checkmark  & \centering\checkmark & \centering\checkmark & \textbf{8.06 $\pm$ 0.22} & {7.48 $\pm$ 0.62 }& {6.40 $\pm$ 0.71 }& \textbf{0.97 $\pm$ 0.16} \\
\hline
\end{tabular}
\end{adjustbox}
\vspace{-0.5cm}
\end{table}

\begin{figure*}[t]
\begin{center}
\includegraphics[width=11.7cm, height=3.4cm]{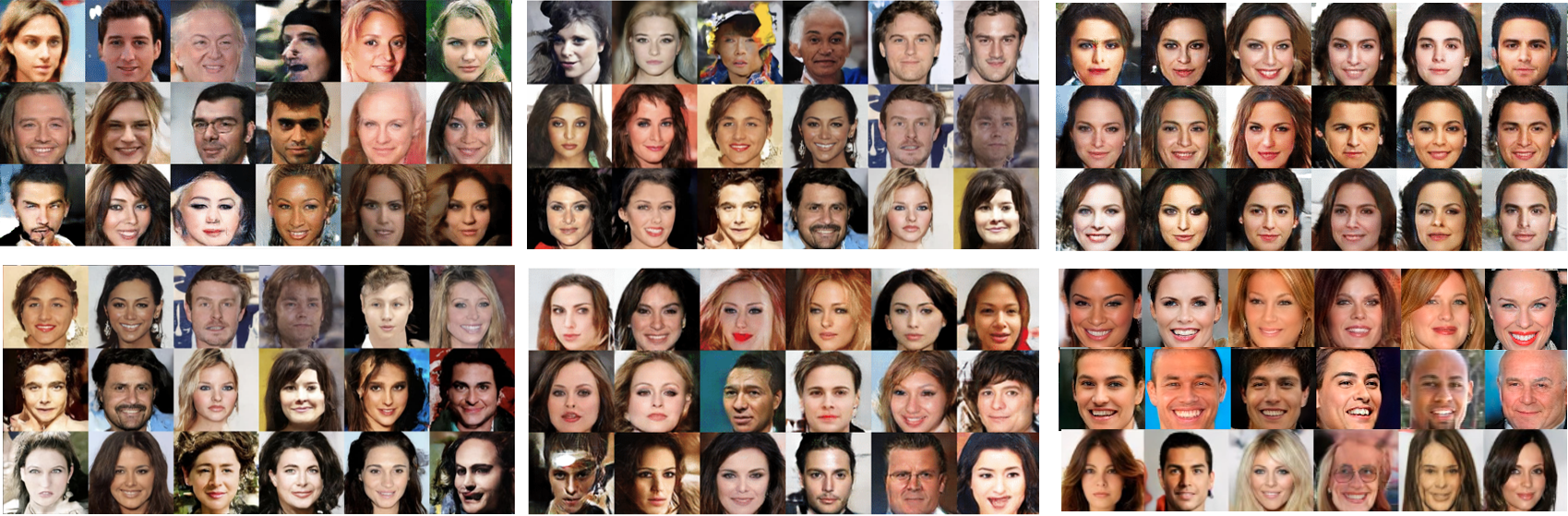}
\end{center}
\vspace{-0.3cm}
\captionlabelfalse 
\caption{(a) \hspace{3cm} (b) \hspace{3cm} (c)}
\vspace{0.3cm}
\captionlabelfalse
   \caption{ Figure 9: Randomly sampled images for QAGAN \cite{Kancharla2019QualityAG} (Top row) and our model (Bottom row) with different losses (SSIM and NIQE) of the CelebA dataset. 
   10 (a) Shows images synthesised using the baseline of both models. 9 (b) The SSIM loss function is added for both cases. We observe similar performance levels. 9 (c) Our model performs better overall with the best results when all loss functions are included ($SSIM+NIQE$).
   }
\label{fig:attribute}
\vspace{0.1cm}
\end{figure*}

\begin{table}[!]
\vspace{0.15cm}
\centering
  \caption{FID on CelebA and CIFAR-10 dataset.}
    \label{table:student}
\begin{adjustbox}{height = 2.4cm, width = 8.2cm}
\begin{tabular}{lc|cc} 
\hline
&\multirow{2}{*} & \multicolumn{1}{c} {\textbf{CIFAR-10 (32 x 32)}} & \multicolumn{1}{c}{\textbf{CelebA (64 x 64)}} \\
\cline{3-4} & \textbf{Model} & \textbf{\textbf{FID}$\downarrow$} & \textbf{\textbf{FID} $\downarrow$} \\
\hline
& QAGAN \cite{Kancharla2019QualityAG}  & 41.20 $\pm$ 0.25 & 10.03 $\pm$ 0.35 \\ 
& QAGAN (SSIM) \cite{Kancharla2019QualityAG} & 14.13 $\pm$ 0.32 & 6.44 $\pm$ 0.43 \\ 
& QAGAN (NIQE) \cite{Kancharla2019QualityAG} & 12.57 $\pm$ 0.11 & 6.40 $\pm$ 0.23 \\ 
& QAGAN (SSIM + NIQE) \cite{Kancharla2019QualityAG}  & 10.01 $\pm$ 0.13 & \textbf{6.16 $\pm$ 0.05}\\
\hline
& StyleGAN2 \cite{Karras2020AnalyzingAI}  & 37.11 $\pm$ 0.15 & 9.03 $\pm$ 0.25 \\ 
& StyleGAN2 (SSIM) \cite{Karras2020AnalyzingAI} & 13.14 $\pm$ 0.02 &\textbf{ 5.84 $\pm$ 0.13} \\
& StyleGAN2 (NIQE) \cite{Karras2020AnalyzingAI} & 11.17 $\pm$ 0.31 & 6.10 $\pm$ 011 \\ 
& StyleGAN2 (SSIM + NIQE) \cite{Karras2020AnalyzingAI}  & 10.81 $\pm$ 0.13 & {6.18 $\pm$ 0.05}\\
\hline
& BL & 37.80 $\pm$ 0.22 & 10.06 $\pm$ 0.43 \\ 
& BL (SSIM) &\textbf{ 12.80 $\pm$ 0.12 }& 6.86 $\pm$ 0.62 \\
& BL (NIQE) & 10.20 $\pm$ 0.79 & 6.36 $\pm$ 0.44 \\
& BL (SSIM + NIQE) &\textbf{9.76 $\pm$ 0.37} & 6.21 $\pm$ 0.36 \\
\hline
\end{tabular}
\end{adjustbox}
\vspace{0.3cm}
\end{table}

\section{Conclusion}
\vspace{-0.3cm}
 In this paper, we introduced a novel quality encoding protocol that harnesses the image quality maps mimicking the HVS and the perceptual properties from a deep convolutional neural network (DCNN) to provide perceptually consistent features that translate to better image quality. We identified  visually sensitive parameters and adapted a quality perceptual attention scheme that narrows down these features to a localised embedding which incentives perceptual representations over other features. The aim was to target the most relevant intrinsic features  responsible for image texture, structural contrast and luminance which we  use to guide the adversarial model towards high quality image synthesis. We also introduced a critic model that monitors perceptual consistency for each image representation. We demonstrated state-of-the-art or comparable performance over other approaches. 

\bibliography{egbib}
\end{document}